\documentclass{article}

\usepackage[main,final]{neurips_2026}

\usepackage[utf8]{inputenc} 
\usepackage[T1]{fontenc}    
\usepackage{hyperref}       
\usepackage{url}            
\usepackage{booktabs}       
\usepackage{amsfonts}       
\usepackage{nicefrac}       
\usepackage{microtype}      
\usepackage[table]{xcolor}  

\usepackage{amsmath}
\usepackage{amssymb}
\usepackage{mathtools}
\usepackage{amsthm}
\usepackage{bbm}
\usepackage{multirow}
\usepackage{mathrsfs}
\usepackage{graphicx}
\usepackage{subfigure}
\usepackage{wrapfig}
\usepackage{epstopdf}
\usepackage{pdfpages}
\usepackage{diagbox}
\usepackage{ulem}
\usepackage{MnSymbol}
\usepackage{pifont} 

\usepackage{tabularx}
\usepackage{makecell}
\usepackage{enumitem}

\newtheorem{theorem}{Theorem}

\DeclareRobustCommand{\ours}{STO}

\newenvironment{tightitemize}{%
  \begin{itemize}[leftmargin=1.4em, labelsep=0.5em, itemsep=0.25em, topsep=0.25em]
}{%
  \end{itemize}
}

\title{Capacity Allocation at the Source: Sparse Target Optimization for LLM Knowledge Editing}

\author{%
  Dongyao Chen\thanks{Equal contribution.} \\
  Renmin University of China \\
  \texttt{dongyaochen@ruc.edu.cn}
  \And
  Jiayun Lei\footnotemark[1] \\
  Renmin University of China \\
  \texttt{leijiayun@ruc.edu.cn}
  \And
  Zhiyuan Ji \\
  Renmin University of China \\
  \texttt{zhiyuanji@ruc.edu.cn}
  \AND
  Zhiying Deng\thanks{Corresponding authors.} \\
  Laboratory for Artificial \\
  Intelligence and New Forms \\
  of Education, Central China \\
  Normal University \\
  \texttt{zhiyingdzy@ccnu.edu.cn}
  \And
    Wei Liu\footnotemark[2] \\
  School of Computing, \\
  National University \\
  of Singapore \\
  \texttt{weiliu87@nus.edu.sg}
  \And
  Xiaobo Sun\footnotemark[2] \\
  Department of Human Genetics, \\
  Emory University \\
  School of Medicine \\
  \texttt{xiaobo.sun@emory.edu}
}

\begin{document}

\normalem
\maketitle

\begin{abstract}
Knowledge editing has emerged as a promising approach for efficiently updating embedded knowledge in large language models (LLMs). It first computes an ideal target hidden state that steers the LLM toward the new output, and then edits the model parameters so that the original hidden state is mapped to this target. While most existing methods focus on the second stage of parameter editing in preserving other knowledge, we find that the upstream construction of the target hidden state also plays an important role in allocating the model's limited capacity space. We first establish a theoretical connection between interference in the parameter space and interference between target hidden states, showing that the two are positively correlated. Building on this insight, we propose a simple yet effective $\ell_1$ regularization method that prunes away unnecessary ``tentacles'' of each knowledge vector, retaining only the essential components. With fewer extraneous tentacles, each edit affects less other knowledge, enabling better coexistence. Extensive experiments across multiple LLMs and benchmark datasets verify the effectiveness of our method. Code: \url{https://github.com/wintersdragon-c/STO}.

\end{abstract}

\section{Introduction}
Large language models (LLMs) encode a substantial amount of factual knowledge in their parameters, but that knowledge is inevitably incomplete, outdated, or erroneous. Updating such knowledge by continued pretraining or full fine-tuning is often computationally expensive and may undesirably perturb unrelated capabilities. This has motivated a growing line of work on model editing, whose goal is to modify a pretrained model so that it incorporates a small set of new facts while preserving unrelated behavior.

\begin{figure}
    \centering
    \includegraphics[width=0.8\linewidth]{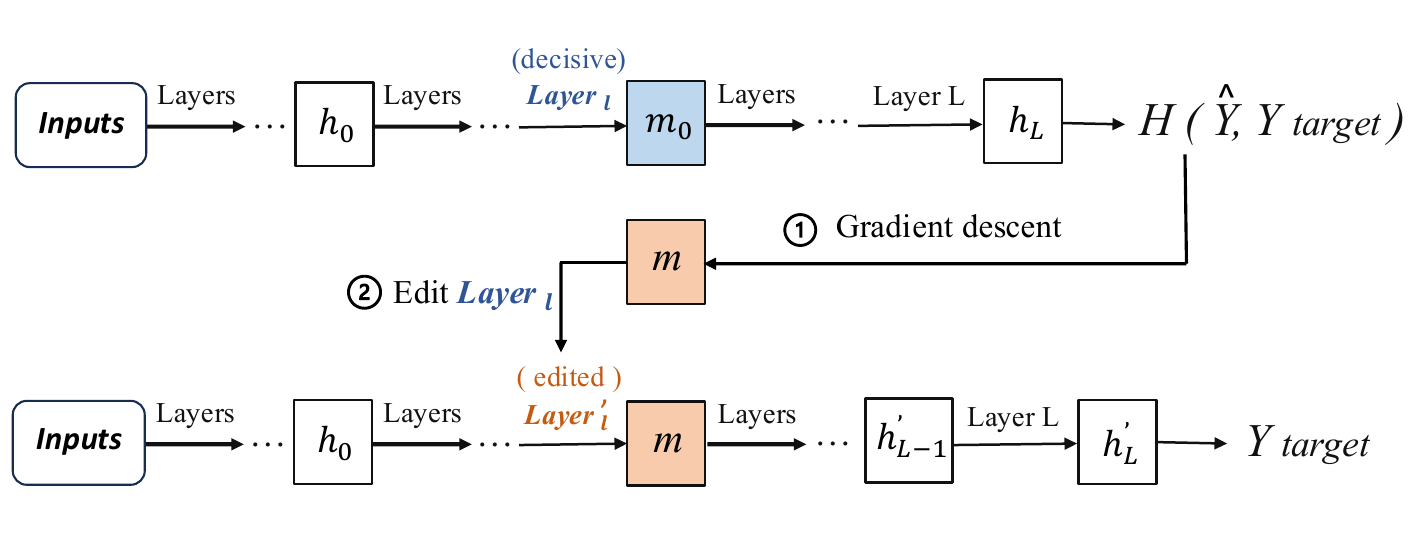}
    \caption{Overview of locate-then-edit model editing. The method first finds a target hidden state for the decisive token at the decisive layer via gradient descent (we call it upstream target construction), then edits a key subset of parameters to steer the original hidden state ($m_0$) toward this target ($m$), producing the desired output ($Y_{target}$) (we call it downstream parameter updating).}
    \label{fig: example of lte}
\end{figure}

One of the mainstream paradigms in this literature is locate-then-edit (LTE) \citep{rome,memit}: First, it precisely locates a decisive token in the input text and identifies the decisive layer in the model (usually the MLP). Then, if the hidden state of the decisive token at the decisive layer can be steered toward a target hidden state (e.g, $m$ in Figure \ref{fig: example of lte}), the model will ultimately output the desired target knowledge.
Figure \ref{fig: example of lte} provides a toy example for quick intuition. 
More details will be introduced in $\S$\ref{sec: preliminaries}.

Despite its strengths, current LTE methods still face a central challenge: because different knowledge items are stored through the same set of parameters, how to better allocate this parameter space and reduce mutual interference among different knowledge, thereby improving the coexistence of more knowledge, remains an under-explored problem. While existing methods account for allocating new knowledge and old non-target knowledge to distinct subspaces when editing parameters based on the residual (i.e., during the second stage of Figure \ref{fig: example of lte}),  the allocation at the upstream target construction stage remains less-explored.  

In this work, we revisit model editing from the perspective of the residual representation (i.e., $m-m_0$ in Figure \ref{fig: example of lte}) that induces the final parameter update and find that we need to consider not only the downstream mechanism of parameter update, but also upstream target construction. 
The capacity space of an LLM is limited. The more new knowledge items that occupy the same region, the higher the interference among them, and the greater the disruption to that region. This not only lowers the success rate of injecting new knowledge into the region but also damages the old knowledge originally stored there. To address this, we attack the problem upstream, at the target construction stage: by spreading new knowledge more thinly across the capacity space, we make space for both new and old knowledge to coexist.

We first theoretically show that the interference between the parameter changes required for editing different pieces of knowledge is positively correlated with the inner product of their residuals. However, due to computational constraints, we cannot compute the residuals for all target knowledge simultaneously when solving via gradient descent, and thus cannot directly adopt inter-residual interference as an optimization regularizer. To address this issue, we try to reduce the expected interference by imposing a simple Lasso \citep{lasso} (i.e., $\ell_1$ regularizer) on the residual vector $r$ (which encourages sparsity in $r$), which can induce a more localized update. Lasso encourages the support of the knowledge representation to be sparse, thereby reducing the expected overlap between the supports of different edits. As a result, the Frobenius interactions among the model parameter updates are also expected to be small. This practical approximation is also quite intuitive: if we limit the residual's variation to only a few necessary dimensions during the residual computation stage, such that the modification of each piece of target knowledge occupies minimal capacity space, would reduce knowledge interference and relieve this allocation burden at its source.

Our contributions can be summarized as follows: (1) We theoretically show that reducing the inner product between the residuals of different knowledge items decreases interference when editing target knowledge. (2) We propose a simple and practical method that reduces the expected inner product between knowledge residuals, thereby lowering editing difficulty from the data preparation stage. (3) Experiments across various LLMs and datasets validate the effectiveness of our method.

\section{Related Work}
\subsection{Locate-then-edit model editing}\label{sec: related lte}
\textbf{Locate-then-edit paradigm.} Large language models (LLMs) encode a substantial amount of factual knowledge in their parameters, but that knowledge is inevitably incomplete, outdated, or erroneous, which motivates concerns on interpretability \citep{liu2025breaking,liu2024mmi}, safety \citep{a2i}, and reliability \citep{liu2026escaping}.
Recent work has increasingly focused on precise parameter editing, typically following a \emph{locate-then-edit} (LTE) paradigm: identifying decisive tokens and layers and then modifying the corresponding hidden states or parameters to substitute target knowledge. Early methods such as ROME and MEMIT \citep{rome,memit} introduced causal tracing to locate effective edit positions.
Subsequent approaches, including RECT \citep{rect}, EMMET \citep{gupta2024unified}, PMET \citep{pmet}, PRUNE \citep{prune}, AdaEdit \citep{adaedit}, AlphaEdit \citep{alphaedit}, LTI \citep{overfit}, AnyEdit \citep{anyedit}, BLUE \citep{residualdistribution}, NeuralEdit \citep{neuraledit}, and FE \citep{FE}, further extend this framework through improved localization, optimization, regularization, or sequential-editing strategies. With strong parameter efficiency and empirical performance, LTE methods have become a major paradigm for knowledge editing \citep{FE}.

\textbf{Target construction of LTE.}
The LTE paradigm consists of two stages: (1) preparing target hidden states, which play a role analogous to labels in supervised learning, (2) using these target states to guide parameter updates. Most prior work focuses on the second stage, namely how to solve or regularize the parameter update, while the upstream construction of target states remains much less explored. MEMIT \citep{memit} argues that the target hidden state should remain close to the original hidden state and introduces an $\ell_2$ distance term to enforce this. EAMET \citep{eamet} further argues that the KL divergence over residuals should match that over keys. Our work also studies target construction, but addresses a different aspect:  we study how target residuals should be allocated within the model's limited capacity space. By sparsifying target residuals, we reduce unnecessary overlap among edits, thereby mitigating interference with other edits and existing non-target knowledge.


\subsection{Other Methods for Knowledge Editing}
\textbf{RAG-style inference-time knowledge editing.}
A parallel line of work maintains an external knowledge base queried at inference time to supply updated information \citep{hartvigsen2023aging, zheng2023can, zhang2024instructedit, jiang2024learning}. Because they leave model parameters unchanged, we regard them as retrieval-based augmentation rather than genuine parameter editing, and they fall outside the scope of this paper.

\textbf{Augmenting models with extra memory or hypernetworks.}
A second family equips LLMs with auxiliary memory modules or hypernetworks to absorb new knowledge. Methods such as MEND \citep{mitchellfast}, KE \citep{de2021editing}, and RLEdit \citep{RLEDIT} exploit low-rank parameterizations for efficiency, yet typically exhibit poor generalization and require retraining for each new fact. Methods that offload knowledge to auxiliary parameters or networks—including T-Patcher \citep{huangtransformer}, SERAC \citep{mitchell2022memory}, Grace \citep{hartvigsen2023aging}, WISE \citep{wise}, and KDE \citep{kde}—face scalability issues as knowledge accumulates without retiring stale entries, and are generally more sensitive to hyperparameter choices than LTE-based approaches.

\section{Preliminaries}\label{sec: preliminaries}

\textbf{Notation}. 
Let $f_{\theta}$ denote a pre-trained LLM parameterized by $\theta$. 
After editing, the updated model is denoted as $f_{\theta^*}$. 
We define the target knowledge set as
\begin{equation}
   S^*=\{(q_i^*, y_i^*)\}_{i=1}^n,
\end{equation}
where $q_i^*$ is an input query that triggers a specific piece of factual knowledge (e.g., \textit{``The president of the US is''}), $y_i^*$ is the desired output after editing (e.g., \textit{``Trump''}), and $n$ denotes the number of knowledge items to be updated.

To encourage the preservation of other knowledge, a representative set
\begin{equation}
   S=\{(q_j, y_j)\}_{j=n+1}^{n+u}
\end{equation}
is typically sampled from a background corpus such as Wikipedia. 
Notably, $y_j$ are
obtained with $y_j=f_\theta(q_j)$, rather than really sampled from Wikipedia.

\textbf{General editing objective}. 
The overall goal of knowledge editing is to update specific factual associations while minimally affecting the model’s behavior on other inputs. 
This objective can be formulated as:
\begin{equation}\label{eqa: overall obj}
\begin{aligned}
    \theta^* = \mathop{\arg\min}_{\hat{\theta}}
    (
    & \sum_{i=1}^n \mathcal{L}_1(f_{\hat{\theta}}(q_i^*), y_i^*) + \lambda \sum_{j=n+1}^{n+u} \mathcal{L}_2(f_{\hat{\theta}}(q_j), y_j)
    ),
\end{aligned}
\end{equation}
where $\mathcal{L}_1$ and $\mathcal{L}_2$ denote loss functions for enforcing the target edit and preserving other knowledge, respectively, and $\lambda$ controls the trade-off between them.

\textbf{Locate-then-edit paradigm}. 
To do editing more efficiently, prior work proposes the \emph{locate-then-edit} (LTE) paradigm \citep{rome,memit} that gives a succinct solution for it.
LTE first identifies where a piece of factual knowledge is stored in the model and then applies targeted parameter updates. A widely adopted approach is \emph{causal tracing} \citep{rome}, which identifies:
\begin{tightitemize}
    \item \textbf{Decisive token}: the token whose hidden representation most strongly determines the factual output, often corresponding to the final token of the subject entity in the input.
    \item \textbf{Decisive layers}: the model layer at which intervening on the hidden state of the decisive token most effectively steers the model toward the desired edit target, often corresponding to several MLP layers.
\end{tightitemize}
By intervening only on the hidden state of the decisive token at the decisive layer, these methods can selectively overwrite specific factual associations with high efficiency and precision. 
Due to its simplicity and strong empirical performance, this paradigm has become one of the mainstreams in the knowledge editing literature.

Despite differences in implementation, many locate-then-edit methods can be expressed as variants of the following optimization problem \citep{memit}:
\begin{equation}\label{eq: memit}
\begin{aligned}
    \Delta^* & = \mathop{\arg\min}_{\Delta}
    (\sum_{i=1}^n \left\| (W + \Delta) k_i - m_i \right\|^2 + \frac{\lambda}{u} \sum_{j=n+1}^{n+u} \left\|  (W + \Delta) k_j - m_j \right\|^2)
    \\
    &=\mathop{\arg\min}_{\Delta}
    (\sum_{i=1}^n \left\| \Delta k_i - (m_i-Wk_i) \right\|^2 + \frac{\lambda}{u} \sum_{j=n+1}^{n+u} \left\| \Delta k_j \right\|^2)
    \\
    W^* &= W + \Delta^*,
\end{aligned}
\end{equation}
where $W$ denotes a selected subset of model parameters to be edited, $k_i$ is the hidden representation of the decisive token at the layer immediately preceding $W$, and $m_i$ is an idealized hidden representation aligned with the desired edit target. The second regularization term is designed to enforce orthogonality between $\Delta$ and other non-target token embeddings $k_j$. The target 
$m_i$ is obtained by treating it as an optimizable parameter and applying gradient descent to minimize the cross-entropy between the current model output and the target output (see Appendix \ref{app: how to get m} for details).

Iteratively solving Eq.~\ref{eq: memit} via gradient descent is computationally expensive, since $u$ is typically very large (e.g., the default is on the order of 40 million). A key advantage of LTE over fine-tuning is that Eq.~\ref{eq: memit} admits a closed-form solution that can be obtained in a single step, efficiently preserving massive non-target knowledge \citep{memit}:
\begin{equation}\label{eqa: memit solution}
    \Delta^*=(M_I-WK_I)K_I^{\top}(K_IK_I^{\top}+\frac{\lambda}{u} K_JK_J^{\top})^{-1},
\end{equation}
where capital letters denote matrices formed by stacking the corresponding vectors. $R=M_I-WK_I$ is the residual for all the knowledge items.

\section{Theoretical Analysis}

LTE methods perform model editing based on Eq. \ref{eq: memit}, which includes two steps. The first step is to obtain the edit target signal $m$ (analogous to a label in classification tasks), and the second step is to solve Eq. \ref{eq: memit} based on this target. Although many existing methods focus on solving Eq. \ref{eq: memit} more effectively, directly obtaining a better $m$ would require much less effort in the subsequent steering process, which is however a less explored avenue.

Here, we only seek a qualitative conclusion to guide the direction of method design. Therefore, we consider a simplified setting for a more detailed theoretical analysis, while still validating our method experimentally in the general setting. Specifically, we consider simultaneously editing two pieces of knowledge, $a$ and $b$, and use the subscript $a/b$ notation to denote the variables associated with these two pieces of knowledge. We denote the edit residual as $r=m_i-Wk_i$. The solution for Eq. \ref{eq: memit} is \citep{memit}
\begin{equation}\label{eq: memit 2 edits}
    \Delta_{a,b}^*=RK_1^\top(\sum_{i\in\{a,b\}} k_ik_i^\top +\frac{\lambda}{u} \sum_j k_jk_j^\top)^{-1},
\end{equation}
where $R=[r_a,r_b], K_1=[k_a,k_b]$. We denote $C_0=\frac{\lambda}{u} \sum_j k_jk_j^\top$.  Typically, $\lambda$ is a very large number. For example, a common choice is $\lambda=15000\gg 2$. Therefore, we usually have $\frac{\lambda}{u} \sum_j k_jk_j^\top\gg \sum_{i\in\{a,b\}} k_ik_i^\top$ and 
we make the following approximation for Eq. \ref{eq: memit 2 edits}: $(\sum_{i\in\{a,b\}} k_ik_i^\top +\frac{\lambda}{u} \sum_j k_jk_j^\top)^{-1}\approx C_0^{-1}$.

Based on the above approximation, we have 
\begin{equation}
    \Delta_{a,b}^*\approx RK_1^\top C_0^{-1}= \sum_{i\in \{a,b\}} r_ik_i^\top C_0^{-1}.
\end{equation}

Similarly, if we consider editing only $a$ or only $b$ separately, we obtain:
\begin{equation}
    \Delta_{a}^*= r_ak_a^\top (k_ak_a^\top+C_0)^{-1}\approx r_ak_a^\top C_0^{-1}, \ \ \ \ \Delta_{b}^*= r_bk_b^\top (k_bk_b^\top+C_0)^{-1}\approx r_bk_b^\top C_0^{-1}.
\end{equation}

Thus, we have 
\begin{equation}
    ||\Delta_{a,b}||_F^2\approx||\Delta_a+\Delta_b||_F^2=||\Delta_{a}||_F^2+||\Delta_{b}||_F^2+2\langle \Delta_{a}, \Delta_{b} \rangle_F, 
\end{equation}
where $\langle \Delta_{a}, \Delta_{b} \rangle_F$ is the Frobenius interaction. If the interaction is large and positive, the two updates will reinforce each other, leading to a larger overall perturbation on other non-target knowledge (i.e., the second term of Eq. \ref{eq: memit}).
If the interaction is large and negative, the two updates $a$ and $b$ will partially cancel each other out, indicating that they are competing in the same parameters (e.g., if $\Delta_a$ and $\Delta_b$ point in exactly opposite directions with identical magnitudes, they will completely cancel each other out).
If the interaction is small, the two updates are more likely to coexist with little mutual interference, and also cause less disruption to the model's existing knowledge.

Based on that, for any two edits, we aim to make their expected interaction $|\langle \Delta_{a}, \Delta_{b} \rangle_F|$ small and thus making the interference of different knowledge edits smaller. 
We denote 
\begin{equation}
    v_a^\top:=k_a^\top (k_ak_a^\top+C_0)^{-1},
\end{equation}
which is irrelevant to $r_a$. Thus we have (note that all equations in the following section are exact and do not involve approximations from the above simplified setting)
\begin{equation}
    \Delta_a=r_av_a^\top, \ \ \ \ \Delta_b=r_bv_b^\top.
\end{equation}
Then, we can get  
\begin{equation}\label{eq: ra*rb}
    \langle \Delta_{a}, \Delta_{b} \rangle_F= \text{tr}(\Delta_a^T\Delta_b)=(r_a^\top r_b)(v_a^\top v_b).
\end{equation}

\begin{theorem}
    If the target residuals of two edits are orthogonal, the corresponding rank-one model updates are Frobenius-orthogonal:
    \begin{equation}
        r_a^\top r_b=0 \Longrightarrow \langle \Delta_{a}, \Delta_{b} \rangle_F=0. 
    \end{equation}
\end{theorem}

Based on the above analysis, to reduce interference between the parameter updates induced by different knowledge edits, we only need to decrease the interference between their residuals $r$, which suggests a new avenue for improving model editing from  the upstream stage of target construction.

\section{A simple and practical solution}
Although our analysis suggests that decreasing the interference between the residuals $r$ of different knowledge edits may be a promising direction, it then leads to the question of how we
can apply this criterion in practice constrained under GPU-memory constraints: different pieces of knowledge are handled independently when computing their corresponding residuals, and during gradient descent to solve for the residual $r$ of a given piece of knowledge, the other knowledge edits are not accessible (due to GPU limitations, it is infeasible to run gradient descent for a large number of knowledge edits simultaneously).

In this study, we present a simple
and practical method.
The overall idea is to make each residual occupy as little space as possible, so that the expected overlap among the residuals of different pieces of knowledge is reduced. Specifically, we use the least absolute shrinkage and selection operator (Lasso) \citep{lasso} to retain only the most essential components of the residual while suppressing trivial patterns. The final method is very simple: when using gradient descent to obtain $m$ in Eq. \ref{eq: memit}, we only need to add an additional Lasso regularization (i.e., $\ell_1$) term:
\begin{equation}
    m^*=\mathop{\arg\min}_{m}(\mathcal{L}_{MEMIT}(m) +\lambda_1||m-Wk||_1),
\end{equation}
where $||m-Wk||_1$ is the added Lasso regularizer, and $\mathcal{L}_{MEMIT}(m)$ is the original basic objective used in MEMIT \citep{memit} (see Appendix \ref{app: loss memit} for details), which typically contains an LLM-related loss (e.g., cross-entropy) and an $\ell_2$ regularizer (i.e., $||m-Wk||_2$).

In terms of Eq.~\ref{eq: ra*rb}, the Lasso regularizer encourages most trivial dimensions in $r_a$ to be near zero and preserves only the components most important for knowledge $a$, thereby reducing the expected overlap between different residuals. This distributes knowledge more sparsely across the LLM's capacity space, both lowering the difficulty of injecting new knowledge and reducing conflicts with existing non-target knowledge. Appendix~\ref{app:sparsity-orthogonality} formalizes the connection between sparsity and orthogonality under a probabilistic support model. 

It is worth noting that aside from the cross-entropy, $\mathcal{L}_{MEMIT}$   already contains an $\ell_2$ distance term (i.e., $||m-Wk||_2$) in addition to the standard cross-entropy. Yet even on top of this existing distance loss, introducing our Lasso loss still yields substantial performance gains. This means that our method's effectiveness does not stem from merely shrinking the magnitude of each dimension ($\ell_2$ alone can already achieve that). Our distinct contribution lies in systematically demonstrating the motivation for introducing Lasso. Diagnostic empirical verification is provided in $\S$\ref{sec: Diagnostic experiments}.


\section{Experiments}


\subsection{Setup}
\textbf{Baselines}. We call our method \textbf{STO} (\textbf{S}parse \textbf{T}arget \textbf{O}ptimization), which is built upon the standard MEMIT framework. MEMIT serves as the direct baseline. We also compare with several recent methods, including , WISE~\citep{wise}, RECT~\citep{rect}, PRUNE~\citep{prune},  AlphaEdit~\citep{alphaedit}, RLEdit \citep{RLEDIT}, EAMET\citep{eamet}.

\textbf{Models, datasets, and metrics.} Experiments are run on an Ascend 910B-64G NPU.
We evaluate STO on three LLMs, which are Qwen2.5-7B-Instruct, GPT-J (6B), and GPT-2 XL (1.5B), following prevailing experimental settings in the model editing literature.
Two benchmark datasets are used: Multi-CounterFact (MCF)~\citep{rome} and ZsRE~\citep{zsre}, each with 2,000 edited knowledge items, following the recent AlphaEdit protocol.
All edits are performed in a single batch (massive editing). For baselines, hyperparameters selection follows their official code or the popular platform EasyEdit~\citep{easyedit}; \ours{} keeps the matched MEMIT editing layers and covariance settings while changing the target-residual construction objective. Appendix~\ref{app:sto-sensitivity} reports the sparsity coefficients, their selection procedure, and sensitivity results.
We report metrics along three standard dimensions.
\textbf{Efficacy} measures whether the target knowledge is correctly injected.
\textbf{Generalization} evaluates robustness under natural linguistic variation, probing whether the model has internalized the knowledge rather than memorized a specific surface form.
\textbf{Specificity} quantifies how well unrelated model behavior is preserved after editing.
Together, these three dimensions capture the core trade-offs between correctness, robustness, and unintended side effects.
For both efficacy and generalization, we report \textbf{Success}~\citep{memit}, a teacher-forcing--based measure, and \textbf{Accuracy}~\citep{editlocality}, which requires exact token match under free-form generation.
For specificity, conventional ground-truth--based metrics have been shown to miss distributional drift induced by editing~\citep{editlocality}.
We therefore adopt the ground-truth--free behavior-deviation metrics proposed by \citet{editlocality}: $D_{KL}$, which measures KL-divergence between pre- and post-edit output distributions, and Top-$k$, which quantifies the overlap ratio of the top-$k$ support.

\subsection{Results on standard benchmarks}

\begin{table}[t]
    \centering
    \caption{The results on the MCF and ZsRE datasets with GPT2-XL.}
    \label{tab:gpt2xl_mcf_zsre_memit_rseq_best}
    \small
    \setlength{\tabcolsep}{2pt}
    \resizebox{\linewidth}{!}{
    \begin{tabular}{@{}ccccccccc@{}}
    \toprule
    \multicolumn{9}{c}{The MCF dataset} \\
    \midrule
    \multicolumn{1}{c}{\multirow{2}{*}{\diagbox{Methods}{Metrics}}} & \multicolumn{2}{c}{Efficacy} & \multicolumn{2}{c}{Generalization} & \multicolumn{4}{c}{Specificity} \\
    \cmidrule(lr){2-3}\cmidrule(lr){4-5}\cmidrule(l){6-9}
    & Success $\uparrow$ & Accuracy $\uparrow$ & Success $\uparrow$ & Accuracy $\uparrow$ & $D_{KL}\downarrow$ & Top-1 $\uparrow$ & Top-5 $\uparrow$ & Top-10 $\uparrow$ \\
    \midrule
    RLedit    & 34.3 & 12.1 & 29.4 & \phantom{0}6.1  & 1.27 & 37.4 & 40.1 & 39.6 \\
    WISE      & 22.0 & \phantom{0}5.0 & 43.0 & 12.5 & 4.95 & 39.0 & 41.0 & 42.0 \\
    MEMIT     & 92.3          & 75.0          & 80.0          & 43.3          & 0.85          & 62.9          & 60.2          & 59.2          \\
    PRUNE     & 92.1          & 75.1          & 79.8          & 42.9          & 0.82          & 63.0          & 60.6          & 59.6          \\
    RECT      & 92.0          & 74.8          & 79.5          & 42.5          & 0.83          & 63.2          & 60.6          & 59.7          \\
    AlphaEdit & 94.8          & 86.8          & 82.2          & 45.8          & 0.98          & 60.5          & 55.8          & 55.0          \\
    EAMET     & 90.8          & 71.6          & 78.3          & 41.2          & 0.84          & 62.3          & 60.4          & 59.3          \\
    \ours{}   & \textbf{95.3} & \textbf{87.6} & \textbf{84.3} & \textbf{46.3} & \textbf{0.76} & \textbf{67.1} & \textbf{63.2} & \textbf{61.5} \\
    \midrule
    \multicolumn{9}{c}{The ZsRE dataset} \\
    \midrule
    \multicolumn{1}{c}{\multirow{2}{*}{\diagbox{Methods}{Metrics}}} & \multicolumn{2}{c}{Efficacy} & \multicolumn{2}{c}{Generalization} & \multicolumn{4}{c}{Specificity} \\
    \cmidrule(lr){2-3}\cmidrule(lr){4-5}\cmidrule(l){6-9}
    & Success $\uparrow$ & Accuracy $\uparrow$ & Success $\uparrow$ & Accuracy $\uparrow$ & $D_{KL}\downarrow$ & Top-1 $\uparrow$ & Top-5 $\uparrow$ & Top-10 $\uparrow$ \\
    \midrule
    RLedit    & 17.2 & \phantom{0}2.1  & \phantom{0}9.5  & \phantom{0}0.3  & 0.99 & 22.0 & 15.4 & 18.4 \\
    WISE      & 13.2 & \phantom{0}1.5  & 15.4 & \phantom{0}0.8  & 2.37 & \phantom{0}6.5  & \phantom{0}4.2  & \phantom{0}9.5  \\
    MEMIT     & 85.0          & 61.2          & 80.7          & 52.2          & 0.81          & 65.7          & 62.4          & 63.9          \\
    PRUNE     & 82.6          & 61.0          & 80.5          & 52.3          & 0.80          & 65.9          & 62.0          & 64.2          \\
    RECT      & 81.0          & 60.1          & 79.0          & 52.2          & 0.79          & 65.8          & 62.4          & 64.0          \\
    AlphaEdit & 93.0          & 82.9          & 88.1          & 68.3          & 0.93          & 67.0          & 63.4          & 64.1          \\
    EAMET     & 86.1          & 60.3          & 79.1          & 51.9          & 0.85          & 66.1          & 62.3          & 65.0          \\
    \ours{}   & \textbf{93.1} & \textbf{84.6} & \textbf{89.6} & \textbf{69.4} & \textbf{0.61} & \textbf{71.6} & \textbf{68.2} & \textbf{69.5} \\
    \bottomrule
    \end{tabular}
    }
    \end{table}

\begin{table}[t]
\centering
\caption{The results on the MCF and ZsRE datasets with Qwen2.5-7B-Instruct.}
\label{tab:qwen25_mcf_zsre_memit_rseq_best}
\small
\setlength{\tabcolsep}{2pt}
\resizebox{\linewidth}{!}{
\begin{tabular}{@{}ccccccccc@{}}
\toprule
\multicolumn{9}{c}{The MCF dataset} \\
\midrule
\multicolumn{1}{c}{\multirow{2}{*}{\diagbox{Methods}{Metrics}}} & \multicolumn{2}{c}{Efficacy} & \multicolumn{2}{c}{Generalization} & \multicolumn{4}{c}{Specificity} \\
\cmidrule(lr){2-3}\cmidrule(lr){4-5}\cmidrule(l){6-9}
& Success $\uparrow$ & Accuracy $\uparrow$ & Success $\uparrow$ & Accuracy $\uparrow$ & $D_{KL}\downarrow$ & Top-1 $\uparrow$ & Top-5 $\uparrow$ & Top-10 $\uparrow$ \\
\midrule
RLedit   & 66.5 & 34.4 & 49.0 & 17.5 & 5.32 & 3.3 & 8.4  & 8.0  \\
WISE     & 28.0 & \phantom{0}6.5 & 53.0 & 16.0 & 6.80 & 25.0 & 27.0 & 27.5 \\
MEMIT     & 99.2          & 92.3          & 97.4          & 70.4          & 2.23          & 41.9          & 41.7          & 38.6          \\
PRUNE     & 99.2          & 92.4          & 97.2          & 70.0          & 2.25          & 41.5          & 41.4          & 39.0          \\
RECT      & 99.1          & 92.2          & 97.0          & 69.7          & 2.24          & 41.6          & 41.5          & 39.1          \\
AlphaEdit & 99.0          & 91.6          & 97.9          & 69.8          & 2.65          & 38.2          & 36.0          & 34.5          \\
EAMET     & 99.1          & 92.0          & 97.3          & 70.0          & 1.98          & 42.3          & 42.0          & 38.9          \\
\ours{}   & \textbf{99.9} & \textbf{99.9} & \textbf{98.0} & \textbf{71.7} & \textbf{0.65} & \textbf{70.0} & \textbf{68.7} & \textbf{67.4} \\
\midrule
\multicolumn{9}{c}{The ZsRE dataset} \\
\midrule
\multicolumn{1}{c}{\multirow{2}{*}{\diagbox{Methods}{Metrics}}} & \multicolumn{2}{c}{Efficacy} & \multicolumn{2}{c}{Generalization} & \multicolumn{4}{c}{Specificity} \\
\cmidrule(lr){2-3}\cmidrule(lr){4-5}\cmidrule(l){6-9}
& Success $\uparrow$ & Accuracy $\uparrow$ & Success $\uparrow$ & Accuracy $\uparrow$ & $D_{KL}\downarrow$ & Top-1 $\uparrow$ & Top-5 $\uparrow$ & Top-10 $\uparrow$ \\
\midrule
RLedit   & 48.7 & 22.3 & 32.4 & \phantom{0}9.0 & 4.20 & 34.6 & \phantom{0}6.1 & \phantom{0}7.0 \\
WISE     & 25.8 & \phantom{0}5.8 & 49.0 & 16.8 & 7.40 & 22.0 & 24.0 & 25.0 \\
MEMIT     & 91.3          & 79.5          & 90.5          & 73.5          & 3.97          & 36.0          & 35.5          & 39.9          \\
PRUNE     & 91.1          & 79.6          & 90.3          & 72.9          & 4.02          & 35.8          & 35.4          & 39.7          \\
RECT      & 90.9          & 79.3          & 90.1          & 72.6          & 4.00          & 35.9          & 35.5          & 39.8          \\
AlphaEdit & 90.7          & 78.8          & 91.0          & 72.8          & 4.35          & 33.0          & 31.0          & 35.6          \\
EAMET     & 94.4          & 85.2          & 92.7          & 79.7          & 3.50          & 40.5          & 39.1          & 42.2          \\
\ours{}   & \textbf{96.6} & \textbf{88.9} & \textbf{95.4} & \textbf{82.6} & \textbf{1.57} & \textbf{63.8} & \textbf{60.0} & \textbf{62.7} \\
\bottomrule
\end{tabular}
}
\end{table}

\begin{figure}[t]
    \centering
    \includegraphics[width=\linewidth]{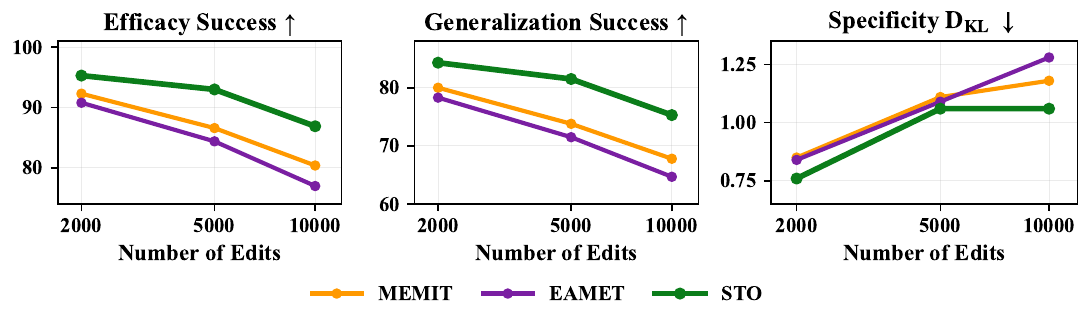}
    \caption{GPT2-XL MCF scaling to 10000 knowledge items.}
    \label{fig:gpt2xl_mcf_scaling_three_metrics}
\end{figure}

The comparison with various recent model editing methods is shown in Table~\ref{tab:gpt2xl_mcf_zsre_memit_rseq_best} and Table~\ref{tab:qwen25_mcf_zsre_memit_rseq_best} (results on GPT-J are in Appendix~\ref{app: additional LLM}). We can observe that our STO achieves the best performance across efficacy, generalization, and specificity. STO can also be combined with PRUNE, RECT, and AlphaEdit by modifying upstream target construction while retaining their downstream solvers. Table~\ref{tab:sto-portability} shows that STO improves overall efficacy, generalization, and specificity across all three frameworks on GPT2-XL; Appendix~\ref{app:sto-integration} provides the corresponding Qwen2.5-7B-Instruct results and integration details. We further scale our method to editing 5000 and 10,000 knowledge items and compare against the most relevant baselines MEMIT and EAMET (see $\S$\ref{sec: related lte} for the relevance). The results are shown in Figure~\ref{fig:gpt2xl_mcf_scaling_three_metrics} and Appendix \ref{app: 10000 edit}. Our advantage becomes even more pronounced at these scales, validating the scalability of our approach. This aligns with our earlier hypothesis: if the interaction is large and positive, the two updates reinforce each other, leading to a larger overall perturbation on non-target knowledge (i.e., lower specificity); if the interaction is large and negative, updates $a$ and $b$ partially cancel each other out, indicating that they compete for the same parameters (i.e., lower efficacy and generalization).

\begin{table}[t]
\centering
\caption{Compatibility of STO with PRUNE, RECT, and AlphaEdit on GPT2-XL (2,000 edits per dataset). Each combination's absolute results are followed by $\Delta$ (combination minus its baseline in Table~\ref{tab:gpt2xl_mcf_zsre_memit_rseq_best}), computed before rounding. Changes are in percentage points except for $D_{KL}$; bold $\Delta$ values indicate improvements.}
\label{tab:sto-portability}
\small
\setlength{\tabcolsep}{2pt}
\resizebox{\linewidth}{!}{
\begin{tabular}{@{}ccccccccc@{}}
\toprule
\multicolumn{9}{c}{The MCF dataset} \\
\midrule
\multicolumn{1}{c}{\multirow{2}{*}{\diagbox{Methods}{Metrics}}} & \multicolumn{2}{c}{Efficacy} & \multicolumn{2}{c}{Generalization} & \multicolumn{4}{c}{Specificity} \\
    \cmidrule(lr){2-3}\cmidrule(lr){4-5}\cmidrule(l){6-9}
    & Success $\uparrow$ & Accuracy $\uparrow$ & Success $\uparrow$ & Accuracy $\uparrow$ & $D_{KL}\downarrow$ & Top-1 $\uparrow$ & Top-5 $\uparrow$ & Top-10 $\uparrow$ \\
    \midrule
PRUNE + STO & 94.4 & 84.8 & 83.0 & 44.2 & 0.70 & 68.5 & 64.4 & 62.7 \\
$\Delta$ & $\mathbf{+2.3}$ & $\mathbf{+9.7}$ & $\mathbf{+3.2}$ & $\mathbf{+1.3}$ & $\mathbf{-0.12}$ & $\mathbf{+5.5}$ & $\mathbf{+3.8}$ & $\mathbf{+3.1}$ \\
\midrule
RECT + STO & 93.6 & 83.4 & 81.8 & 42.1 & 0.68 & 69.1 & 65.0 & 63.3 \\
$\Delta$ & $\mathbf{+1.6}$ & $\mathbf{+8.6}$ & $\mathbf{+2.3}$ & $-0.4$ & $\mathbf{-0.15}$ & $\mathbf{+5.9}$ & $\mathbf{+4.4}$ & $\mathbf{+3.6}$ \\
\midrule
AlphaEdit + STO & 99.7 & 99.2 & 89.9 & 52.9 & 0.49 & 75.4 & 70.8 & 69.0 \\
$\Delta$ & $\mathbf{+4.9}$ & $\mathbf{+12.4}$ & $\mathbf{+7.7}$ & $\mathbf{+7.1}$ & $\mathbf{-0.49}$ & $\mathbf{+14.9}$ & $\mathbf{+15.0}$ & $\mathbf{+14.0}$ \\
\midrule
\multicolumn{9}{c}{The ZsRE dataset} \\
\midrule
\multicolumn{1}{c}{\multirow{2}{*}{\diagbox{Methods}{Metrics}}} & \multicolumn{2}{c}{Efficacy} & \multicolumn{2}{c}{Generalization} & \multicolumn{4}{c}{Specificity} \\
    \cmidrule(lr){2-3}\cmidrule(lr){4-5}\cmidrule(l){6-9}
    & Success $\uparrow$ & Accuracy $\uparrow$ & Success $\uparrow$ & Accuracy $\uparrow$ & $D_{KL}\downarrow$ & Top-1 $\uparrow$ & Top-5 $\uparrow$ & Top-10 $\uparrow$ \\
    \midrule
PRUNE + STO & 93.0 & 83.3 & 88.5 & 66.6 & 0.41 & 76.7 & 73.4 & 74.8 \\
$\Delta$ & $\mathbf{+10.4}$ & $\mathbf{+22.3}$ & $\mathbf{+8.0}$ & $\mathbf{+14.3}$ & $\mathbf{-0.40}$ & $\mathbf{+10.8}$ & $\mathbf{+11.4}$ & $\mathbf{+10.6}$ \\
\midrule
RECT + STO & 92.9 & 82.6 & 88.2 & 65.8 & 0.40 & 76.7 & 73.4 & 74.8 \\
$\Delta$ & $\mathbf{+11.9}$ & $\mathbf{+22.5}$ & $\mathbf{+9.2}$ & $\mathbf{+13.6}$ & $\mathbf{-0.39}$ & $\mathbf{+10.9}$ & $\mathbf{+11.0}$ & $\mathbf{+10.8}$ \\
\midrule
AlphaEdit + STO & 97.5 & 98.7 & 95.2 & 80.2 & 0.47 & 73.5 & 70.0 & 71.3 \\
$\Delta$ & $\mathbf{+4.5}$ & $\mathbf{+15.8}$ & $\mathbf{+7.1}$ & $\mathbf{+11.9}$ & $\mathbf{-0.46}$ & $\mathbf{+6.5}$ & $\mathbf{+6.6}$ & $\mathbf{+7.2}$ \\
\bottomrule
\end{tabular}
}
\end{table}

We next design more fine-grained experiments to validate our method in greater detail.

\begin{figure}[t]
    \centering
    \renewcommand{\subcapsize}{\small\bfseries}
    \subfigure[Near-zero fraction across thresholds]{
        \includegraphics[width=0.42\linewidth]
        {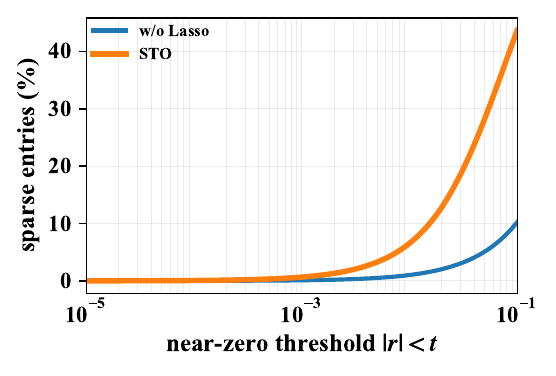}
    }
    \hfill
    \subfigure[Residual magnitude density]{
        \includegraphics[width=0.42\linewidth]
        {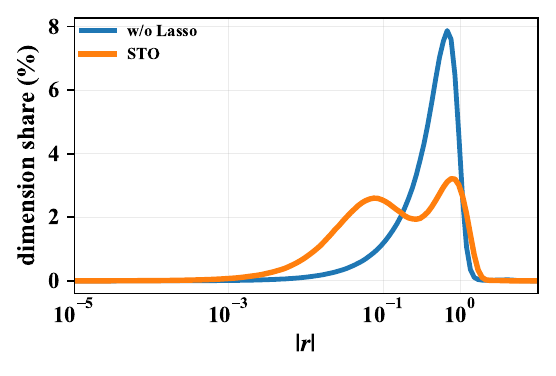}
    }

    \caption{Verifying feature sparsification. (a) We use varying thresholds (x-axis) to define near-zero values; the y-axis shows the proportion of dimensions falling below each threshold. (b) Distribution of per-dimension values: the x-axis denotes the value of each dimension in the residual, and the y-axis denotes density.}
    \label{fig:2}
\end{figure}

\subsection{Diagnostic experiments}\label{sec: Diagnostic experiments}

To better validate our method, we design targeted experiments addressing several specific questions. 

\textbf{Question (1)}: Can the Lasso regularizer really lead to sparse residuals? 

\begin{figure}[t]
    \centering
    \renewcommand{\subcapsize}{\small\bfseries}
    \subfigure[Representative groups: high, middle, and low]{
        \includegraphics[width=0.45\linewidth]
        {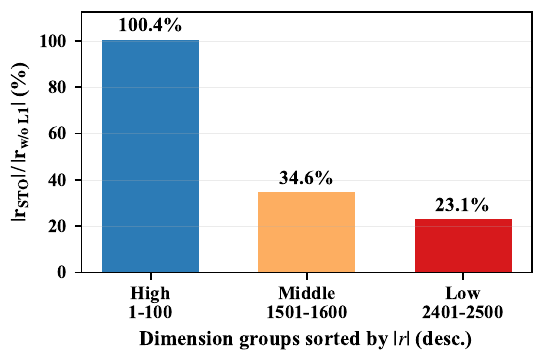}
    }
    \hfill
    \subfigure[All dimension groups]{
        \includegraphics[width=0.45\linewidth]
        {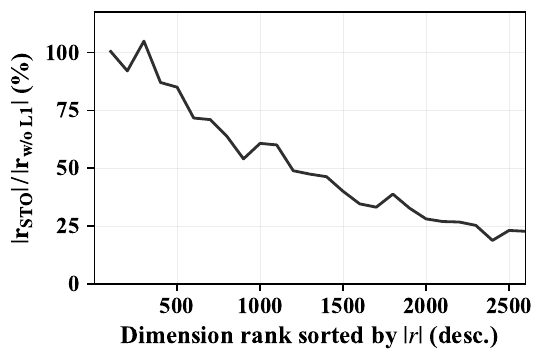}
    }
    \caption{Dimension-level $|\frac{r_{sto}}{r_{w/o\  \ell_1}}|$. We sort the dimensions of $r$ in descending order by absolute value and compute $|\frac{r_{sto}}{r_{w/o\  \ell_1}}|$. For ease of visualization, we group every 100 adjacent dimensions and report the average ratio per group.}
    \label{fig:3}
\end{figure}

We compare the residuals obtained by our method against those obtained without the Lasso constraint, measuring the proportion of dimensions whose values are close to zero. We use several different thresholds to determine whether a dimension is considered near-zero. The results are shown in Figure~\ref{fig:2}.
Figure~\ref{fig:2}(a) shows that our method substantially increases sparsity. Figure~\ref{fig:2}(b) plots the sample density (y-axis) of residual values across different magnitude ranges (x-axis). STO's density function clearly exhibits two peaks: Compared to the no-Lasso baseline, the rightmost curve almost coincides with the original, while a second peak emerges around 0.1, indicating that certain unimportant dimensions have indeed been compressed while the important features are preserved. Our goal is to preserve important features while removing unnecessary ones. Is the phenomenon in Figure \ref{fig:2}(a) due to simple universal weight shrinkage?
To further verify this, we sort the dimensions of residual vector \(r\) by magnitude and aggregate them in groups of 100, then examine the compression ratio across dimensions with and without Lasso. The results are shown in Figures~\ref{fig:3}(a) and \ref{fig:3}(b). Figure~\ref{fig:3}(a) reports the compression ratio for the low, middle, and high regions, while Figure~\ref{fig:3}(b) shows the compression ratio across all dimensions continuously. This indicates that our method does not simply compress all dimensions; rather, it largely preserves the originally high-magnitude, important dimensions (retaining approximately $100.4\%$)  while more aggressively compressing the low-magnitude, unimportant ones. This aligns with the design principle of Lasso \citep{lasso} that it does feature selection rather than simple weight shrinkage.

\textbf{Question (2)}: How much does our method affect the original cross-entropy loss? 

Cross-entropy measures how accurately a given piece of knowledge is steered toward the target. Therefore, to ensure successful downstream editing, our method should not significantly affect the cross-entropy. We select 2,000 samples from MCF dataset and conduct experiments using Qwen2.5-7B-Instruct. We report the mean and median cross-entropy values, as well as the residual inner product corresponding to Eq.~\ref{eq: ra*rb} (computed over 500K sampled pairs).

As shown in Table~\ref{tab:residual_interaction_diagnostics}, the inner product between residuals is significantly reduced, while the cross-entropy increases only marginally. Combined with the finding from Question (1)—that our method preserves high-magnitude dimensions and primarily compresses low-magnitude ones—this confirms that we retain the necessary dimensions while suppressing unimportant ones. 

Notably, although a higher cross-entropy would in principle suggest a disadvantage in editing success rate (Efficacy), our method instead achieves higher Efficacy, particularly on the harder ZsRE dataset. This indicates that reducing inter-knowledge interference indeed improves editing success.

\textbf{Question (3)}: Does the performance gain simply come from reducing the magnitude of each dimension in the residual $r$? 

The comparison between our method and MEMIT directly addresses this concern. Our approach can be viewed as augmenting MEMIT with an additional Lasso constraint ($\ell_1$) on top of its existing $\ell_2$ regularization on the residual. These two regularizers play complementary roles: $\ell_2$ performs weight shrinkage by uniformly attenuating parameter magnitudes, while $\ell_1$ induces sparsity and thereby acts as an implicit feature selector, retaining informative components while pruning redundant ones. The performance gain of our method over MEMIT therefore indicates that feature selection, beyond mere weight shrinkage, is essential in this setting. In practice, naively strengthening the $\ell_2$ term in MEMIT does not resolve the issue and in fact hurts performance, since $\ell_2$ penalizes all features indiscriminately and ends up suppressing the informative ones together with the redundant ones.

\begin{table}[t]
\centering
\caption{We report the mean/average cross-entropy (CE) loss and sample 500K  $(a,b)$ pairs to report $|r_a^\top r_b|$. We achieve much lower knowledge interference (Eq. \ref{eq: ra*rb}) but only marginally higher CE. }
\label{tab:residual_interaction_diagnostics}
\small
\setlength{\tabcolsep}{5pt}
\resizebox{0.7\linewidth}{!}{
\begin{tabular}{@{}lcccc@{}}
\toprule
  Method & \(|r_a^\top r_b|\) (mean)$\downarrow$  &  \(|r_a^\top r_b|\) (median)$\downarrow$  &CE (mean) &CE (median) \\
\midrule
  \ours{}        & 251.2152 & 195.9598 & 0.0161 & 0.0021 \\
  w/o $\ell_1$   & 541.1298 & 442.3562 & 0.0152 & 0.0020 \\

\bottomrule
\end{tabular}
}
\end{table}

\section{Conclusion, limitations, and future work}

This paper analyzes, from the perspective of target construction, how to better allocate target knowledge within the limited capacity space of LLMs during editing. We first establish a theoretical connection between interference in the parameter space and interference between target residuals, showing that the two are positively correlated. Building on this insight, we propose STO, a simple yet effective $\ell_1$ regularization method that prunes away unnecessary $``$branches'' of each knowledge vector, leaving only the essential components. With fewer extraneous branches, knowledge items are less likely to entangle, and the interference between residuals (and consequently in the LLM parameter space) is substantially reduced.
Our results suggest that target construction is also a fruitful direction for improving the reliability of model editing. We hope this work encourages further investigation into how the structure of edit targets, not only the mechanism of parameter updates, shapes the success of LLM model editing.  One limitation is that we do not have results for LLMs larger than 14B, mainly due to two practical reasons. One reason is that the hyperparameters for most baselines are often unavailable for larger LLMs. In addition, larger-scale experiments exceed our current computational resources. Nevertheless, the settings we use are commonly used in the literature, and we leave the study of larger LLMs to future work.

\begin{ack}
This work is supported by the National Natural Science Foundation of China under Grant 62507018 and the Hubei Provincial Natural Science Foundation of China under Grant JCZRMS202601468. It is also  supported by the National Research Foundation, Singapore, and the Ministry of Digital Development
and Information (MDDI) under the Singapore Global AI Visiting Professorship Program (Award No. AIVP-2024-002). 
We also thank the AFAC (Advanced Fintech AI Competition) Academic Acceleration Program for supporting this research.
The authors declare no competing interests.
\end{ack}

\bibliography{iclr}
\bibliographystyle{iclr}

\clearpage

\appendix

\begin{table}[h]
\centering
\caption{The results on the MCF and ZsRE datasets with GPT-J.}
\label{tab:gptj_mcf2000_memit_rseq_best}
\small
\setlength{\tabcolsep}{2pt}
\resizebox{\linewidth}{!}{
\begin{tabular}{@{}ccccccccc@{}}
\toprule
\multicolumn{9}{c}{The MCF dataset} \\
\midrule
\multicolumn{1}{c}{\multirow{2}{*}{\diagbox{Methods}{Metrics}}} & \multicolumn{2}{c}{Efficacy} & \multicolumn{2}{c}{Generalization} & \multicolumn{4}{c}{Specificity} \\
\cmidrule(lr){2-3}\cmidrule(lr){4-5}\cmidrule(l){6-9}
& Success $\uparrow$ & Accuracy $\uparrow$ & Success $\uparrow$ & Accuracy $\uparrow$ & $D_{KL}\downarrow$ & Top-1 $\uparrow$ & Top-5 $\uparrow$ & Top-10 $\uparrow$ \\
\midrule
RLedit    & 77.6 & 52.0 & 45.3 & 18.7 & 2.97 & \phantom{0}6.7 & 13.7 & 16.2 \\
WISE      & 27.5 & \phantom{0}7.0 & 51.0 & 15.5 & 4.50 & 45.8 & 48.0 & 49.5 \\
MEMIT     & 99.7          & 99.4          & 92.6          & 66.9          & 0.55          & 72.6          & 69.4          & 69.1          \\
PRUNE     & 99.7          & 99.4          & 92.7          & 66.6          & 0.52          & 73.0          & 70.0          & 69.8          \\
RECT      & 99.8          & 99.3          & 92.4          & 66.0          & 0.53          & 72.9          & 70.1          & 69.9          \\
AlphaEdit & 99.7          & 98.8          & 94.0          & 66.2          & 0.67          & 70.2          & 64.4          & 63.6          \\
EAMET     & 99.7          & 98.8          & 91.5          & 67.0          & 0.63          & 70.4          & 67.8          & 67.6          \\
\ours{}   & \textbf{99.9} & \textbf{99.5} & \textbf{95.7} & \textbf{67.8} & \textbf{0.48} & \textbf{76.6} & \textbf{73.4} & \textbf{72.8} \\
\midrule
\multicolumn{9}{c}{The ZsRE dataset} \\
\midrule
\multicolumn{1}{c}{\multirow{2}{*}{\diagbox{Methods}{Metrics}}} & \multicolumn{2}{c}{Efficacy} & \multicolumn{2}{c}{Generalization} & \multicolumn{4}{c}{Specificity} \\
\cmidrule(lr){2-3}\cmidrule(lr){4-5}\cmidrule(l){6-9}
& Success $\uparrow$ & Accuracy $\uparrow$ & Success $\uparrow$ & Accuracy $\uparrow$ & $D_{KL}\downarrow$ & Top-1 $\uparrow$ & Top-5 $\uparrow$ & Top-10 $\uparrow$ \\
\midrule
RLedit    & 88.1 & 72.7 & 73.3 & 54.8 & 0.41 & 75.6 & 71.4 & 72.5 \\
WISE      & 50.7 & 19.2 & 48.5 & 21.0 & 11.01 & \phantom{0}1.8 & \phantom{0}2.1 & \phantom{0}2.2 \\
MEMIT     & 99.1          & 96.6          & 94.6          & 92.5          & 0.23          & 82.1          & 74.3          & 76.0          \\
PRUNE     & \textbf{99.2} & \textbf{96.8}          & 94.5          & 92.4          & 0.22          & 82.2          & 74.5          & 76.1          \\
RECT      & 99.0          & 96.5          & 94.4          & 92.3          & 0.23          & 82.3          & 74.6          & 76.0          \\
AlphaEdit & 85.5          & 72.2          & 74.0          & 59.7          & 0.46          & 80.1          & 72.3          & 74.5          \\
EAMET     & 90.4          & 85.3          & 78.9          & 64.6          & 0.19          & 87.4          & 79.6          & 81.7          \\
\ours{}   & 96.8 & 89.1 & \textbf{94.7} & \textbf{92.8} & \textbf{0.18} & \textbf{87.6} & \textbf{79.9} & \textbf{82.5} \\
\bottomrule
\end{tabular}
}
\end{table}

\begin{table}[t]
    \centering
    \caption{\textbf{Qwen2.5-7B-Instruct 10000-edit comparison.}The MCF block reports completed runs on on the same exact 10,000-edit manifest.}
    \label{tab:qwen25_mcf10000_memit_rseq}
    \small
    \setlength{\tabcolsep}{2pt}
    \resizebox{\linewidth}{!}{
    \begin{tabular}{@{}lrrrrrrrr@{}}
    \toprule
    \multicolumn{9}{c}{The MCF dataset} \\
    \midrule
    \multirow{2}{*}{\diagbox{Methods}{Metrics}} & \multicolumn{2}{c}{Efficacy} & \multicolumn{2}{c}{Generalization} & \multicolumn{4}{c}{Specificity} \\
    \cmidrule(lr){2-3}\cmidrule(lr){4-5}\cmidrule(l){6-9}
    & Success $\uparrow$ & Accuracy $\uparrow$ & Success $\uparrow$ & Accuracy $\uparrow$ & $D_{KL}\downarrow$ & Top-1 $\uparrow$ & Top-5 $\uparrow$ & Top-10 $\uparrow$ \\
    \midrule
  MEMIT      & 73.7 & 5.7  & 62.7 & 2.0  & 5.69 & 8.2  & 12.5 & 12.3 \\
  EAMET     & 74.9 & 10.2  &   65.1   & 3.5      &  5.20   &  9.0   &  12.4   &   10.3            \\
    \ours{}  & \textbf{91.6} & \textbf{59.0} & \textbf{86.7} & \textbf{32.0} & \textbf{4.72} & \textbf{19.7} & \textbf{20.5} & \textbf{18.6} \\
    \midrule
    \end{tabular}
    }
\end{table}

 \begin{table}[t]
  \centering
  \caption{\textbf{GPT2-XL 10000-edit comparison.}The MCF block reports completed runs on on the same exact 10,000-edit manifest.}
  \label{tab:gpt2xl_mcf10000_memit_rseq_best}
  \small
  \setlength{\tabcolsep}{2pt}
  \resizebox{\linewidth}{!}{
  \begin{tabular}{@{}lrrrrrrrr@{}}
  \toprule
  \multicolumn{9}{c}{The MCF dataset} \\
  \midrule
  \multirow{2}{*}{\diagbox{Methods}{Metrics}} & \multicolumn{2}{c}{Efficacy} & \multicolumn{2}{c}{Generalization} &
  \multicolumn{4}{c}{Specificity} \\
  \cmidrule(lr){2-3}\cmidrule(lr){4-5}\cmidrule(l){6-9}
  & Success $\uparrow$ & Accuracy $\uparrow$ & Success $\uparrow$ & Accuracy $\uparrow$ & $D_{KL}\downarrow$ & Top-1 $
  \uparrow$ & Top-5 $\uparrow$ & Top-10 $\uparrow$ \\
  \midrule
  MEMIT      & 80.4&47.6&67.8& 26.6&1.18&53.4&52.0&51.2      \\
  EAMET      & 77.0   & 43.4   & 64.7  & 25.1   & 1.28   & 52.2  & 51.4  & 50.1   \\
  \ours{}    & \textbf{86.9}   & \textbf{61.4}  & \textbf{75.3}   & \textbf{31.9}  & \textbf{1.06}   & \textbf{56.8}   & \textbf{55.4}   &  \textbf{53.1}   \\
  \bottomrule
  \end{tabular}
  }
  \end{table}
  

\section{How to get the target hidden state $m$} \label{app: how to get m}

In Eq.~\ref{eq: memit}, a critical step is to obtain the target hidden state $m$.
Intuitively, $m$ plays the role of an intermediate target representation: if the
model can be made to produce this representation at the selected token and layer,
then the remaining layers should assign high probability to the desired new
answer. The standard MEMIT-style approach treats this hidden state as an
optimizable variable and solves for it by gradient descent, using a loss that
encourages the model to predict the target answer. We describe the MEMIT loss in
Appendix~\ref{app: loss memit}.

Concretely, suppose the rewrite query is tokenized as
$Q=[q_1,q_2,\ldots,q_d,\ldots,q_n]$, where $q_d$ denotes the decisive token, and
the desired answer is tokenized as $Y=[y_1,y_2,y_3]$. Let layer-$l$ be the final
decisive layer. After feeding the concatenated sequence $[Q,Y]$ through the
first $l$ layers of the LLM, we obtain hidden states
$h(Q,Y)=[h_{q_1},\ldots,h_{q_d},\ldots,h_{q_n},h_{y_1},\ldots,h_{y_3}]$.

We denote the remaining layers of the model by $\hat{Y}=f(h(Q,Y))$. The
cross-entropy loss is computed only on the target answer tokens, and is denoted
by $H_c(Y,\hat{Y})$. During target-state optimization, the model parameters are
frozen, and the hidden representation of the decisive token, $h_{q_d}$, is
treated as the optimizable variable. Minimizing $H_c(Y,\hat{Y})$ with respect to
this hidden state gives an optimized representation, which is taken as the
target hidden state $m$.

Prior work shows that, if the selected MLP parameters are updated so that the
original representation at the decisive token is mapped to $m$, the remaining
layers of the model are encouraged to generate the target answer
\citep{rome,memit}. The preservation term in Eq.~\ref{eq: memit} further
regularizes the update so that non-target key representations are affected as
little as possible.

\section{The loss function of MEMIT} \label{app: loss memit}

This appendix summarizes the target-state optimization loss used in MEMIT
\citep{memit}. Before applying its closed-form parameter update, MEMIT first
constructs a target hidden state $m_i$ for each edit request. This target state
specifies what the representation of the decisive token should become so that
the frozen remaining layers of the model generate the desired new answer.

Given an edit request $(s_i,r_i,o_i^*)$, let $p(s_i,r_i)$ denote the rewrite
prompt, and let $h_i^L$ be the hidden state of the last subject token at the
target layer $L$. MEMIT parameterizes the candidate target state as
$m_i=h_i^L+\delta_i$, where $\delta_i$ is an optimizable residual. During this
target-construction step, all model parameters are frozen; only $\delta_i$ is
updated.

The main objective is to make the model assign high probability to the desired
object $o_i^*$ when the decisive-token hidden state is replaced by $m_i$:
\begin{equation}
\mathcal{L}_{\mathrm{CE}}(m_i)
=
\frac{1}{P}
\sum_{j=1}^{P}
-\log
P_{G(h_i^L \leftarrow m_i)}
\left[
o_i^*
\mid
x_j \oplus p(s_i,r_i)
\right],
\end{equation}
where $G(h_i^L \leftarrow m_i)$ denotes a forward pass in which the original
hidden state $h_i^L$ is replaced by the candidate target state $m_i$. The random
prefixes $\{x_j\}_{j=1}^{P}$ make the optimized target state robust to different
surface forms of the prompt, rather than overfitting to a single prompt.

MEMIT also keeps the optimized target close to the model's original
representation. In the notation of our main paper, let $m_{0,i}=Wk_i$ denote the
pre-edit value representation produced by the selected MLP module, and let
$r_i=m_i-m_{0,i}$ denote the target residual. The MEMIT target-state loss can be
written as
\begin{equation}
\mathcal{L}_{\mathrm{MEMIT}}(m_i)
=
\mathcal{L}_{\mathrm{CE}}(m_i)
+
\lambda_2 \|m_i-m_{0,i}\|_2^2 .
\end{equation}
The first term encourages the edited model to predict the new object, while the
$\ell_2$ term discourages the target state from moving too far from the original
representation.

Our method keeps this MEMIT objective and adds an $\ell_1$ regularizer on the
same residual, giving
$\mathcal{L}_{\mathrm{STO}}(m_i)
=
\mathcal{L}_{\mathrm{MEMIT}}(m_i)
+
\lambda_1\|m_i-m_{0,i}\|_1$.
Equivalently, STO optimizes
$\mathcal{L}_{\mathrm{CE}}(m_i)
+\lambda_2\|m_i-m_{0,i}\|_2^2
+\lambda_1\|m_i-m_{0,i}\|_1$.
Thus, the original MEMIT loss finds a target hidden state that induces the
desired answer while remaining close to the original representation; our added
$\ell_1$ term further encourages the residual $r_i=m_i-m_{0,i}$ to use only a
smaller set of essential dimensions before the standard MEMIT closed-form update
is applied.

\section{Scaling to 10000 knowledge items}\label{app: 10000 edit}

We further evaluate whether \ours{} remains effective as the number of
simultaneous edits increases. Tables~\ref{tab:qwen25_mcf10000_memit_rseq}
and~\ref{tab:gpt2xl_mcf10000_memit_rseq_best} report the 10,000-edit MCF
results on Qwen2.5-7B-Instruct and GPT2-XL. 


\section{Results on additional LLM}\label{app: additional LLM}
Corresponding to Table \ref{tab:gpt2xl_mcf_zsre_memit_rseq_best} and Table \ref{tab:qwen25_mcf_zsre_memit_rseq_best}, the results of the GPT-J model are in Table \ref{tab:gptj_mcf2000_memit_rseq_best}.


\section{Compatibility with additional LTE frameworks}
\label{app:sto-integration}

STO acts on upstream target-residual construction. To combine it with PRUNE, RECT, or AlphaEdit, we add the $\ell_1$ penalty when optimizing each target residual and then pass the resulting targets to the framework's existing downstream solver. Each combination retains the framework's downstream procedure.

Table~\ref{tab:qwen-sto-integration} reports the Qwen2.5-7B-Instruct compatibility results on MCF and ZsRE, with 2,000 edits per dataset. Each combination is shown with its absolute results and changes from the corresponding baseline across all eight metrics; the GPT2-XL results are in Table~\ref{tab:sto-portability} in the main text. On Qwen2.5-7B-Instruct, all three frameworks obtain higher efficacy and better specificity after adding STO. Most generalization metrics also improve, although some MCF metrics decrease slightly.

\begin{table}[htbp]
\centering
\caption{Compatibility of STO with PRUNE, RECT, and AlphaEdit on Qwen2.5-7B-Instruct (2,000 edits per dataset). Each combination's absolute results are followed by $\Delta$ (combination minus its baseline in Table~\ref{tab:qwen25_mcf_zsre_memit_rseq_best}), computed before rounding. Changes are in percentage points except for $D_{KL}$; bold $\Delta$ values indicate improvements.}
\label{tab:qwen-sto-integration}
\small
\setlength{\tabcolsep}{2pt}
\resizebox{\linewidth}{!}{
\begin{tabular}{@{}ccccccccc@{}}
\toprule
\multicolumn{9}{c}{The MCF dataset} \\
\midrule
\multicolumn{1}{c}{\multirow{2}{*}{\diagbox{Methods}{Metrics}}} & \multicolumn{2}{c}{Efficacy} & \multicolumn{2}{c}{Generalization} & \multicolumn{4}{c}{Specificity} \\
    \cmidrule(lr){2-3}\cmidrule(lr){4-5}\cmidrule(l){6-9}
    & Success $\uparrow$ & Accuracy $\uparrow$ & Success $\uparrow$ & Accuracy $\uparrow$ & $D_{KL}\downarrow$ & Top-1 $\uparrow$ & Top-5 $\uparrow$ & Top-10 $\uparrow$ \\
    \midrule
PRUNE + STO & 99.7 & 99.5 & 97.6 & 69.5 & 0.68 & 69.1 & 68.6 & 67.9 \\
$\Delta$ & $\mathbf{+0.5}$ & $\mathbf{+7.1}$ & $\mathbf{+0.4}$ & $-0.5$ & $\mathbf{-1.57}$ & $\mathbf{+27.6}$ & $\mathbf{+27.2}$ & $\mathbf{+28.9}$ \\
\midrule
RECT + STO & 99.7 & 99.5 & 97.5 & 69.4 & 0.67 & 68.9 & 68.3 & 67.7 \\
$\Delta$ & $\mathbf{+0.6}$ & $\mathbf{+7.3}$ & $\mathbf{+0.5}$ & $-0.3$ & $\mathbf{-1.57}$ & $\mathbf{+27.3}$ & $\mathbf{+26.8}$ & $\mathbf{+28.6}$ \\
\midrule
AlphaEdit + STO & 99.8 & 99.7 & 97.8 & 71.8 & 0.58 & 71.1 & 69.7 & 68.8 \\
$\Delta$ & $\mathbf{+0.8}$ & $\mathbf{+8.1}$ & $-0.1$ & $\mathbf{+2.0}$ & $\mathbf{-2.07}$ & $\mathbf{+32.9}$ & $\mathbf{+33.7}$ & $\mathbf{+34.3}$ \\
\midrule
\multicolumn{9}{c}{The ZsRE dataset} \\
\midrule
\multicolumn{1}{c}{\multirow{2}{*}{\diagbox{Methods}{Metrics}}} & \multicolumn{2}{c}{Efficacy} & \multicolumn{2}{c}{Generalization} & \multicolumn{4}{c}{Specificity} \\
    \cmidrule(lr){2-3}\cmidrule(lr){4-5}\cmidrule(l){6-9}
    & Success $\uparrow$ & Accuracy $\uparrow$ & Success $\uparrow$ & Accuracy $\uparrow$ & $D_{KL}\downarrow$ & Top-1 $\uparrow$ & Top-5 $\uparrow$ & Top-10 $\uparrow$ \\
    \midrule
PRUNE + STO & 96.8 & 97.8 & 95.6 & 93.7 & 1.55 & 61.1 & 64.0 & 65.1 \\
$\Delta$ & $\mathbf{+5.7}$ & $\mathbf{+18.2}$ & $\mathbf{+5.3}$ & $\mathbf{+20.8}$ & $\mathbf{-2.47}$ & $\mathbf{+25.3}$ & $\mathbf{+28.6}$ & $\mathbf{+25.4}$ \\
\midrule
RECT + STO & 96.8 & 97.9 & 95.7 & 94.0 & 1.56 & 60.8 & 63.6 & 64.6 \\
$\Delta$ & $\mathbf{+5.9}$ & $\mathbf{+18.6}$ & $\mathbf{+5.6}$ & $\mathbf{+21.4}$ & $\mathbf{-2.44}$ & $\mathbf{+24.9}$ & $\mathbf{+28.1}$ & $\mathbf{+24.8}$ \\
\midrule
AlphaEdit + STO & 96.7 & 92.2 & 95.3 & 84.1 & 1.81 & 63.4 & 66.2 & 68.0 \\
$\Delta$ & $\mathbf{+6.0}$ & $\mathbf{+13.4}$ & $\mathbf{+4.3}$ & $\mathbf{+11.3}$ & $\mathbf{-2.54}$ & $\mathbf{+30.4}$ & $\mathbf{+35.2}$ & $\mathbf{+32.4}$ \\
\bottomrule
\end{tabular}
}
\end{table}

\section{Sparsity coefficient and sensitivity analysis}
\label{app:sto-sensitivity}

STO retains the original MEMIT target objective, including its $\ell_2$ term, and adds the residual penalty $\lambda_1\|m-Wk\|_1$. Table~\ref{tab:sto-coefficients} lists the coefficients used for the standard MEMIT + STO experiments. We selected them by monitoring optimization dynamics in TensorBoard and manually balancing the loss terms. Apart from $\lambda_1$, the sensitivity comparisons use the corresponding MEMIT hyperparameters from EasyEdit.

\begin{table}[htbp]
\centering
\caption{Selected sparsity coefficients for the standard MEMIT + STO experiments.}
\label{tab:sto-coefficients}
\small
\begin{tabular}{@{}llc@{}}
\toprule
Model & Dataset & $\lambda_1$ \\
\midrule
GPT2-XL & MCF & $9\times10^{-4}$ \\
GPT2-XL & ZsRE & $6\times10^{-4}$ \\
Qwen2.5-7B-Instruct & MCF / ZsRE & $3\times10^{-4}$ \\
GPT-J-6B & MCF / ZsRE & $3\times10^{-4}$ \\
\bottomrule
\end{tabular}
\end{table}

We vary $\lambda_1$ over $\{0,3\times10^{-4},6\times10^{-4},9\times10^{-4}\}$ on GPT2-XL and Qwen2.5-7B-Instruct, with 2,000 edits per dataset. The zero-penalty runs provide the matched MEMIT baselines for this sweep. Table~\ref{tab:sto-sensitivity} reports efficacy, generalization, and specificity together.

Every tested positive coefficient improves efficacy accuracy and Top-1 preservation over its corresponding zero-penalty baseline. Increasing the positive coefficient reduces $D_{KL}$ within each setting, but can reduce efficacy or generalization. For example, on GPT2-XL / MCF, increasing $\lambda_1$ from $3\times10^{-4}$ to $9\times10^{-4}$ reduces $D_{KL}$ from 0.97 to 0.76 while generalization accuracy changes from 60.0 to 46.3. The coefficient therefore controls an editing--preservation trade-off, and the selected operating point need not maximize every metric.

\begin{table}[htbp]
\centering
\caption{Sensitivity to the sparsity coefficient with 2,000 edits per setting. Coefficients in the first column are in units of $10^{-4}$; bold coefficients mark the selected operating points.}
\label{tab:sto-sensitivity}
\small
\setlength{\tabcolsep}{6pt}
\begin{tabular}{@{}rrrrrrr@{}}
\toprule
$\lambda_1$ & Eff-S $\uparrow$ & Eff-A $\uparrow$ & Gen-S $\uparrow$ & Gen-A $\uparrow$ & $D_{KL}\downarrow$ & Top-1 $\uparrow$ \\
\midrule
\multicolumn{7}{c}{GPT2-XL / MCF} \\
\midrule
0 & 92.3 & 75.0 & 80.0 & 43.3 & 0.85 & 62.9 \\
3 & 98.6 & 95.8 & 92.2 & 60.0 & 0.97 & 63.0 \\
6 & 97.8 & 93.6 & 89.2 & 53.7 & 0.85 & 65.2 \\
\textbf{9} & 95.3 & 87.6 & 84.3 & 46.3 & 0.76 & 67.1 \\
\midrule
\multicolumn{7}{c}{GPT2-XL / ZsRE} \\
\midrule
0 & 85.0 & 61.2 & 80.7 & 52.2 & 0.81 & 65.7 \\
3 & 95.2 & 90.8 & 92.0 & 74.7 & 0.62 & 70.9 \\
\textbf{6} & 93.1 & 84.6 & 89.6 & 69.4 & 0.61 & 71.6 \\
9 & 88.8 & 72.0 & 83.7 & 56.3 & 0.38 & 77.8 \\
\midrule
\multicolumn{7}{c}{Qwen2.5-7B-Instruct / MCF} \\
\midrule
0 & 99.2 & 92.3 & 97.4 & 70.4 & 2.23 & 41.9 \\
\textbf{3} & 99.9 & 99.9 & 98.0 & 71.7 & 0.65 & 70.0 \\
6 & 99.6 & 99.4 & 96.0 & 66.5 & 0.62 & 78.7 \\
9 & 98.6 & 98.3 & 93.7 & 62.0 & 0.47 & 82.8 \\
\midrule
\multicolumn{7}{c}{Qwen2.5-7B-Instruct / ZsRE} \\
\midrule
0 & 91.3 & 79.5 & 90.5 & 73.5 & 3.97 & 36.0 \\
\textbf{3} & 96.6 & 88.9 & 95.4 & 82.6 & 1.57 & 63.8 \\
6 & 96.6 & 97.7 & 94.2 & 90.7 & 1.07 & 74.1 \\
9 & 96.7 & 99.1 & 94.1 & 92.3 & 0.89 & 79.4 \\
\bottomrule
\end{tabular}
\end{table}

\section{Probabilistic connection between sparsity and orthogonality}
\label{app:sparsity-orthogonality}

This appendix formalizes the support-overlap motivation for STO. Probability is taken over random edit residuals in a fixed model and coordinate system. For $r_a,r_b\in\mathbb{R}^d$, define their nonzero supports as $T_i=\{t:r_{i,t}\neq0\}$ for $i\in\{a,b\}$. Let $X=|T_a\cap T_b|$ and $q_{ab,t}=\Pr(t\in T_a\cap T_b)$.

\begin{theorem}[Support-overlap bound]
For arbitrary joint distributions of the two supports,
\begin{equation}
\label{eq:support-orthogonality-bound}
\Pr(r_a^\top r_b=0)
\geq \Pr(X=0)
\geq \max\left\{0,1-\sum_{t=1}^{d}q_{ab,t}\right\}.
\end{equation}
\end{theorem}

\begin{proof}
Writing $X=\sum_{t=1}^{d}\mathbf{1}\{t\in T_a\cap T_b\}$ gives $\mathbb{E}[X]=\sum_t q_{ab,t}$ by linearity of expectation. Since $X$ is a nonnegative integer, Markov's inequality implies $\Pr(X\geq1)\leq\mathbb{E}[X]$. Hence $\Pr(X=0)\geq1-\mathbb{E}[X]$, together with the trivial lower bound of zero. Disjoint supports imply $r_a^\top r_b=0$, which proves the first inequality. Orthogonality may also occur through cancellation when the supports overlap, so disjointness is sufficient but not necessary.
\end{proof}

To recover a product form, assume that, for each coordinate $t$, the support indicators of the two edits are independent. Defining $\pi_{i,t}=\Pr(t\in T_i)$ then gives
\begin{equation}
\label{eq:support-overlap-product}
\mathbb{E}[X]=\sum_{t=1}^{d}\pi_{a,t}\pi_{b,t},
\qquad
\Pr(r_a^\top r_b=0)\geq
\max\left\{0,1-\sum_{t=1}^{d}\pi_{a,t}\pi_{b,t}\right\}.
\end{equation}
Independence between different coordinates is not required. Under this model, decreasing the coordinate-usage probabilities can improve the lower bound. More generally, the relevant quantity is the expected joint overlap $\sum_t q_{ab,t}$.

An analogous magnitude bound connects overlap to the residual factor in Eq.~\ref{eq: ra*rb}. If $|r_{a,t}|\leq B_a$ and $|r_{b,t}|\leq B_b$ almost surely for every coordinate, then
\begin{equation}
\label{eq:support-magnitude-bound}
\mathbb{E}\!\left[|r_a^\top r_b|\right]
\leq B_aB_b\,\mathbb{E}[X]
=B_aB_b\sum_t q_{ab,t}.
\end{equation}
This follows by bounding the absolute inner product by $\sum_t|r_{a,t}r_{b,t}|\leq B_aB_bX$. For fixed key-side factors $v_a,v_b$, Eq.~\ref{eq: ra*rb} multiplies this bound by $|v_a^\top v_b|$ for the corresponding rank-one update interaction.

\paragraph{Scope of the argument.}
These statements provide a conditional motivation for sparsity, not a guarantee that an $\ell_1$ penalty makes residuals orthogonal. A smaller support or $\ell_1$ norm alone need not reduce joint overlap: two sparse residuals can concentrate on the same coordinates. The probability bound is vacuous when expected overlap is at least one, and a larger lower bound does not establish monotonicity of the actual orthogonality probability. Finite-step target optimization typically produces near-zero coefficients rather than exact zeros, so exact supports are an idealization. The empirical inner-product and cross-entropy comparison in Table~\ref{tab:residual_interaction_diagnostics} tests the resulting residual geometry directly. Finally, Frobenius orthogonality of updates is a geometric statement and does not by itself guarantee preservation of all non-target model behavior.

\section{Application note}
Our subsequent AFAC competition entry on financial data factor development cites this published work and uses STO as a technical prototype for further development.

\clearpage
\section*{NeurIPS Paper Checklist}

\begin{enumerate}

\item {\bf Claims}
    \item[] Question: Do the main claims made in the abstract and introduction accurately reflect the paper's contributions and scope?
    \item[] Answer: \answerYes{}
    \item[] Justification: The abstract and Introduction state the paper's scope: a target-construction method for locate-then-edit model editing, a residual-interference analysis, and experiments on multiple LLMs and editing benchmarks. The claims are supported by the theoretical analysis, standard benchmark tables, diagnostic experiments, and scaling appendix.

\item {\bf Limitations}
    \item[] Question: Does the paper discuss the limitations of the work performed by the authors?
    \item[] Answer: \answerYes{}
    \item[] Justification: The paper states the method's scope in the Conclusion: One limitation is that we do not have results for LLMs larger than 14B, mainly due to two practical reasons. One reason is that the hyperparameters for most baselines are often unavailable for larger LLMs. In addition, larger-scale experiments exceed our current computational resources. Nevertheless, the settings we use are commonly used in the literature, and we leave the study of larger LLMs to future work.

\item {\bf Theory assumptions and proofs}
    \item[] Question: For each theoretical result, does the paper provide the full set of assumptions and a complete (and correct) proof?
    \item[] Answer: \answerYes{}
    \item[] Justification: The theoretical analysis states the two-edit setting, residual notation, covariance approximation used for intuition, and the exact residual-interaction identity in Eq.~\ref{eq: ra*rb}. The theorem follows directly from that identity and is cross-referenced with the method motivation and diagnostics.

    \item {\bf Experimental result reproducibility}
    \item[] Question: Does the paper fully disclose all the information needed to reproduce the main experimental results of the paper to the extent that it affects the main claims and/or conclusions of the paper (regardless of whether the code and data are provided or not)?
    \item[] Answer: \answerYes{}
    \item[] Justification: The paper specifies the models, datasets, metrics, edit setting, target-construction objective, and benchmark protocol in the Experiments section and appendices. Code, configurations, and running instructions are provided at \url{https://github.com/wintersdragon-c/STO}.

\item {\bf Open access to data and code}
    \item[] Question: Does the paper provide open access to the data and code, with sufficient instructions to faithfully reproduce the main experimental results, as described in supplemental material?
    \item[] Answer: \answerYes{}
    \item[] Justification: The public repository \url{https://github.com/wintersdragon-c/STO} provides code, configuration files, and running instructions.

\item {\bf Experimental setting/details}
    \item[] Question: Does the paper specify all the training and test details (e.g., data splits, hyperparameters, how they were chosen, type of optimizer) necessary to understand the results?
    \item[] Answer: \answerYes{}
    \item[] Justification: The Experiments section describes the evaluated models, MCF and ZsRE datasets, massive-editing setup, baselines, metrics, and matched MEMIT settings. Additional target-state optimization and loss details are provided in the appendices and artifact repository.

\item {\bf Experiment statistical significance}
    \item[] Question: Does the paper report error bars suitably and correctly defined or other appropriate information about the statistical significance of the experiments?
    \item[] Answer: \answerNo{}
    \item[] Justification: The main experiments are controlled matched-manifest comparisons: all methods are evaluated on the same edit sets with the same evaluation pipeline, and the paper does not claim random-seed or hyperparameter-search statistical significance. We do not report error bars or confidence intervals because this is not standard in the model-editing benchmark protocol; the release artifacts include manifests and metric-generation code for auditing or extending the comparisons.

\item {\bf Experiments compute resources}
    \item[] Question: For each experiment, does the paper provide sufficient information on the computer resources (type of compute workers, memory, time of execution) needed to reproduce the experiments?
    \item[] Answer: \answerYes{}
    \item[] Justification: The release artifacts document the compute setup used to reproduce the experiments, including accelerator requirements, model/dataset preparation, and commands for target construction, editing, and evaluation. The paper also discusses scalability through the 10,000-edit appendix experiments.
    
\item {\bf Code of ethics}
    \item[] Question: Does the research conducted in the paper conform, in every respect, with the NeurIPS Code of Ethics \url{https://neurips.cc/public/EthicsGuidelines}?
    \item[] Answer: \answerYes{}
    \item[] Justification: This research is limited to theoretical analysis and offline benchmark evaluation of a methodological improvement for locate-then-edit knowledge editing. It uses existing benchmark datasets and pretrained language models, and does not involve human-subject experiments, private or sensitive data collection, user-facing deployment, or the release of newly trained foundation models. The artifact release is limited to research code, configurations, and reproduction instructions, and does not include edited model checkpoints. To the best of our knowledge, the work conforms to the NeurIPS Code of Ethics.

\item {\bf Broader impacts}
    \item[] Question: Does the paper discuss both potential positive societal impacts and negative societal impacts of the work performed?
    \item[] Answer: \answerNA{}
    \item[] Justification: This work is a methodological study of capacity allocation in locate-then-edit knowledge editing. It provides theoretical analysis and offline benchmark experiments on how sparse target construction affects edit interference. The paper does not introduce a deployed system, collect or release new human data, train or release a new foundation model, or provide edited model checkpoints. Therefore, we do not identify direct societal impacts or specific harmful downstream consequences beyond the general research context of model-editing methods.
    
\item {\bf Safeguards}
    \item[] Question: Does the paper describe safeguards that have been put in place for responsible release of data or models that have a high risk for misuse (e.g., pre-trained language models, image generators, or scraped datasets)?
    \item[] Answer: \answerYes{}
    \item[] Justification: The paper does not release a new pretrained model, an edited model checkpoint, or a new high-risk dataset. The released artifact is limited to public research code, configurations, and reproduction instructions for offline benchmark evaluation. Access to the underlying pretrained models and benchmark datasets remains governed by their original licenses and usage terms. These release boundaries reduce the risk that the artifact directly enables deployment of altered models while still supporting scientific reproducibility.

\item {\bf Licenses for existing assets}
    \item[] Question: Are the creators or original owners of assets (e.g., code, data, models), used in the paper, properly credited and are the license and terms of use explicitly mentioned and properly respected?
    \item[] Answer: \answerYes{}
    \item[] Justification: The paper cites the benchmark datasets, pretrained model families, editing baselines, and public editing frameworks used in the experiments. The public artifact repository documents external dependencies and points users to the original asset licenses and access terms.

\item {\bf New assets}
    \item[] Question: Are new assets introduced in the paper well documented and is the documentation provided alongside the assets?
    \item[] Answer: \answerYes{}
    \item[] Justification: The new release asset is the public code and reproduction package at \url{https://github.com/wintersdragon-c/STO}. It documents the method implementation, configurations, evaluation scripts, and reproduction workflow for offline benchmark evaluation.

\item {\bf Crowdsourcing and research with human subjects}
    \item[] Question: For crowdsourcing experiments and research with human subjects, does the paper include the full text of instructions given to participants and screenshots, if applicable, as well as details about compensation (if any)? 
    \item[] Answer: \answerNA{}
    \item[] Justification: The paper does not involve crowdsourcing, user studies, participant interaction, or human-subject experiments. All experiments are offline evaluations on existing model-editing benchmarks.

\item {\bf Institutional review board (IRB) approvals or equivalent for research with human subjects}
    \item[] Question: Does the paper describe potential risks incurred by study participants, whether such risks were disclosed to the subjects, and whether Institutional Review Board (IRB) approvals (or an equivalent approval/review based on the requirements of your country or institution) were obtained?
    \item[] Answer: \answerNA{}
    \item[] Justification: No human subjects are involved, so IRB or equivalent human-subjects review is not applicable. The study uses existing public benchmarks and offline model evaluations.

\item {\bf Declaration of LLM usage}
    \item[] Question: Does the paper describe the usage of LLMs if it is an important, original, or non-standard component of the core methods in this research? Note that if the LLM is used only for writing, editing, or formatting purposes and does \emph{not} impact the core methodology, scientific rigor, or originality of the research, declaration is not required.
    \item[] Answer: \answerYes{}
    \item[] Justification: LLMs are the subject of study: the paper edits pretrained language models and evaluates their behavior before and after editing. The method does not rely on a separate LLM as an undisclosed component for data generation, labeling, or evaluation beyond the documented model-editing experiments.

\end{enumerate}

\end{document}